\documentclass[final]{cvpr}

\usepackage{times}
\usepackage{epsfig}
\usepackage{graphicx}
\usepackage{amsmath}
\usepackage{amssymb}
\usepackage{cuted}
\usepackage{capt-of}

\usepackage{enumitem}
\setitemize{noitemsep,topsep=0pt,parsep=0pt,partopsep=0pt}
\setenumerate{noitemsep,topsep=0pt,parsep=0pt,partopsep=0pt}

\DeclareMathOperator*{\argmin}{argmin}

\usepackage[pagebackref=true,breaklinks=true,colorlinks,bookmarks=false]{hyperref}

\newcommand{\trainingIdentities}{$58$}
\newcommand{\testIdentities}{$6$}
\newcommand{\totalIdentities}{$64$}

\begin{document}
\title{i3DMM: Deep Implicit 3D Morphable Model of Human Heads}

\author{
Tarun Yenamandra\textsuperscript{1}, Ayush Tewari\textsuperscript{2}, Florian Bernard\textsuperscript{1}, Hans-Peter Seidel\textsuperscript{2},
Mohamed Elgharib\textsuperscript{2}, \\ Daniel Cremers\textsuperscript{1}, and Christian Theobalt\textsuperscript{2} \\
\textsuperscript{1}TU Munich, \textsuperscript{2}MPI Informatics, Saarland Informatics Campus
}
\maketitle
\begin{strip}
    \includegraphics[width=\linewidth]{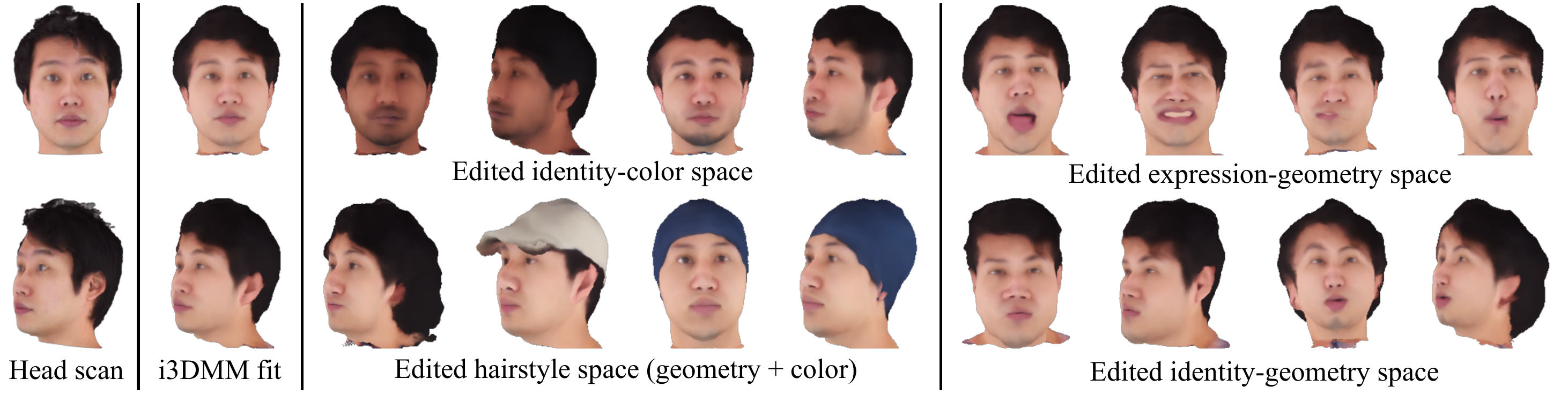}
    \captionof{figure}{
    Our deep implicit 3D morphable model 
    (i3DMM) of human heads includes semantically disentangled spaces for color (identity and hairstyle), and geometry (identity, expression, and hairstyle). 
    We can fit i3DMM to head scans, and edit the different components.
    } 
    \label{fig:teaser}
\end{strip}
\begin{abstract} \vspace{-10pt}
We present the first deep implicit 3D morphable model (i3DMM) of full heads. 
Unlike earlier morphable face models it not only captures identity-specific geometry, texture, and expressions of the frontal face, but also models the entire head, including hair.
We collect a new dataset consisting of~\totalIdentities{} people with different expressions and hairstyles to train i3DMM. 
Our approach has the following favorable properties: (i) It is the first full head morphable model that includes hair. (ii) In contrast to mesh-based models it can be trained on merely rigidly aligned scans, without requiring difficult non-rigid registration. (iii) 
We design a novel architecture to decouple the shape model into an implicit reference shape and a deformation of this reference shape. With that, dense correspondences between shapes can be learned implicitly.
(iv) This architecture allows us to  semantically disentangle the geometry and color components, as color is learned in the reference space. 
Geometry is further disentangled as identity, expressions, and hairstyle, while color is disentangled as identity and hairstyle components. 
We show the merits of i3DMM using ablation studies, comparisons to state-of-the-art models, and applications such as semantic head editing and texture transfer. We will make our model publicly available.
\end{abstract}

\section{Introduction}
3D morphable models (3DMMs) are parametric models of geometry and appearance of human faces, with widespread use in applications such as image and video editing, face recognition and cognitive science~\cite{egger20193d}.
These models are trained using 3D scans of humans, e.g., laser scans~\cite{Blanz1999}, depth sensor-based scans~\cite{Cao2014b}, or photometric multi-view scans~\cite{Li17}.
As 3DMMs usually learn a deformation space of a fixed template mesh, the training shapes commonly need to be brought into dense surface correspondence with each other. 
Computing such correspondence for the face region alone is already hard and often requires a challenging non-convex optimizaton problem~\cite{egger20193d}; computing dense correspondence for the rest of the head and the hair is close to impossible. 
Therefore, as well as due to the lack of large 3D scan datasets with hair, most 3DMMs only capture the face region. 
While some recent approaches aim to model the complete head~\cite{Dai2019,Li17,Ranjan18,ploumpis2019combining}, 
they do not capture the hair region, and the appearance in the head region (if captured) is very limited. 

In this work, we present i3DMM, the first implicit 3D morphable model which captures the full head region, including hair deformations and appearance. 
We capture   
a new dataset of full photogrammetric head scans of~\totalIdentities{} people for training. 
Each subject performs several expressions and natural hair is captured. 
Since such scans are noisy, especially in the hair region,
our training algorithm uses an adaptive sampling strategy based on the quality of reconstructions in different regions of the head, allowing us to effectively handle noise without smoothing out details.  
In contrast to existing 3DMMs, which use a mesh-representation with a fixed template, we implicitly represent surfaces using signed distance functions. This allows us to capture large deformations easily, which is particularly convenient for the hair region.
The implicit representation also allows us to avoid computing dense correspondence between training scans.
Our method is inspired by recent works on deep implicit surface modeling~\cite{Park_2019_CVPR,Mescheder19Occupancy,NeRF}, which demonstrate the advantages of using an implicit representation compared to voxel grids, point clouds or meshes.  
Our main technical innovation compared to these works is that we use a novel neural network architecture which \emph{decouples the learning process}, separating it into learning a reference shape, learning geometry deformations with respect to this reference, and learning a color network. 
This allows us to automatically compute dense correspondences between any two shapes with only sparse supervision on salient face landmarks. %
This decoupling is also essential for disentangling the learned space into semantically meaningful parametric components, an important feature of 3DMMs. 

Most classical models disentangle identity geometry, expression geometry, and appearance of the face. 
This is sufficient for several applications in face editing~\cite{ThiesZSTN2016a,tewari2020pie,sun2018hybrid}, but does not allow the explicit control of head parts other than the facial region.
Our model allows for unprecedented control over the different semantic modes of full heads. 
To this end, we 
learn to disentangle the geometry and color, 
where geometry is further separated into identity, expression, and hairstyle components, and color is further separated into identity and hairstyle components. 

In summary, our main contributions are:
\begin{enumerate}
    \item A method for learning full head 3D morphable models directly from rigidly aligned real-world noisy 3D scans without dense ground truth correspondences.
    \item A novel network architecture which can compute dense correspondences between 3D shapes represented using implicit functions.
    This network is trained only with sparse supervision. 
    \item A training method for disentangling
    \begin{enumerate}
        \item the color and geometry components, and
        \item the identity, expression, and hairstyle components of the geometry; and the identity and hairstyle components of the color.
    \end{enumerate}
\end{enumerate}
We compare our model to several state-of-the-art 3DMMs by fitting to 3D scans. We also demonstrate the quality of our learned dense correspondences by showing texture transfer between scans, and show that i3DMM benefits applications such as semantic head editing, and one-shot computation of segmentation masks and landmarks on 3D scans.

\section{Related Work}
Most approaches for face and head modeling are mesh-based. We summarize them here and refer the reader to Egger~\etal~\cite{egger20193d} for a more comprehensive report. As our model is based on implicit representations, we also discuss methods which learn implicit shape and color.%

\textbf{Face and Head Morphable Models.}
The work of Blanz and Vetter~\cite{BlanzVetter} showed the possibility of representing faces using a 3D Morphable Model (3DMM).
The model is learned by transforming the shape and texture of examples into low-dimensional vector representations using principal components analysis (PCA).
Further improvements were proposed~\cite{Gerig2017MorphableFM,Paysan09,Booth16,Booth18} by using better registration and higher quality scans. %
Multilinear models have also been proposed to model identity-dependent expressions~\cite{FWCao14,Abrevaya18}.
\newline\indent
Li~\etal~\cite{FLAME:SiggraphAsia2017} presented FLAME,
a head model that combines a linear shape space with an articulated jaw, neck and eyeballs, pose-dependent corrective expression blendshapes, and global expressions.
The model is learned from a large 3D dataset, including data from D3DFACS~\cite{Cosker11}.
Ranjan~\etal~\cite{COMA:ECCV18} learned a head model using an autoencoder based on spectral graph convolutions~\cite{Defferrard16}.
The LYHM head model~\cite{Dai19,Dai17} uses a hierarchical parts-based template morphing framework to process the head shape, and uses optical flow to refine the texture. 
The model is built using $1{,}200$ different identities.
Ploumpis~\etal~\cite{ploumpis2019combining,ploumpis2019towards} combined the LYHM and LSFM~\cite{booth20163d} models as the Universal Head model (UHM), with a focus on face geometry.
Note that all existing full head models only model the cranium geometry without hair.

\textbf{Implicit Representation.} Implicit representations have a long history in computer vision for inferring 3D shape \cite{Leventon-et-al-00}, temporally evolving shape \cite{Cremers-06,Kohlberger-et-al-06} and textured shape \cite{Sturm-et-al-gcpr13}. Park~\etal~\cite{Park_2019_CVPR} presented DeepSDF to represent a class of shapes using signed distance fields with an autodecoder.
Chen~\etal~\cite{chen2018implicit_decoder} and Mescheder~\etal~\cite{Mescheder19Occupancy} learned a generative model of a class of shapes by classifying a point in space as inside or outside a shape.
{Recent approaches~\cite{Tretschk2020PatchNets,deng2020cvxnet,genova2020local} have proposed to  represent shapes using a collection of local implicit patches for higher quality.
}
Oechsle~\etal~\cite{Oechsle2019ICCV} extended OccupancyNets~\cite{Mescheder19Occupancy} to represent texture along with the geometry using monocular inputs.
Niemeyer~\etal~\cite{Niemeyer2019ICCV} presented an approach for handling time-varying shapes by learning to predict motion vectors for each point in 3D space using OccupancyNets.
PIFu~\cite{saito2019pifu,saito2020pifuhd} allows for 3D reconstruction of humans from monocular inputs. 
{Pixel-aligned implicit functions estimate a continuous field that determines whether a pixel is inside or outside the surface of the human subject, as well as the color on the surface}. 
 Several recent methods~\cite{sitzmann2019siren,NeRF,tancik2020fourfeat} have presented ways to achieve higher-quality implicit representations by using periodic functions as activations.
Saito~\etal~\cite{Saito18} presented an approach for 3D hair modeling. 
Their technique learns a manifold for 3D hairstyles represented as implicit surfaces using a volumetric variational autoencoder.

\section{Method}
\begin{figure*}[ht!]
\includegraphics[width=\linewidth]{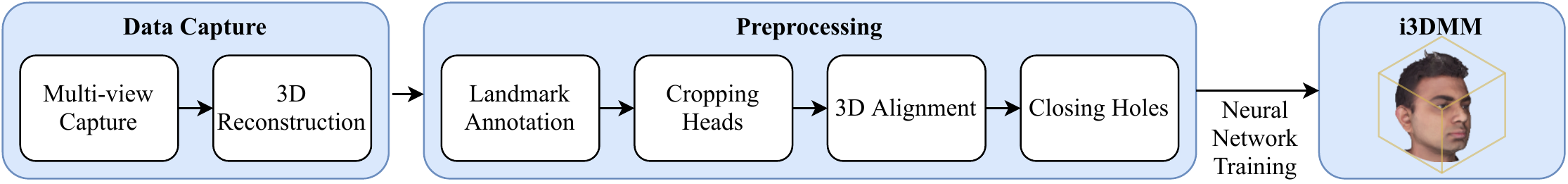}
    \caption{Overview of our approach.
    }
    \label{fig:pipeline} 
\end{figure*}
In this section, we describe how we obtain our deep implicit 3D morphable model (i3DMM). Our overall processing pipeline is illustrated in Fig.~\ref{fig:pipeline}. In the following we will describe our data acquisition, the training data preparation, and the neural network architecture.
\begin{figure}[t]
\includegraphics[width=\linewidth]{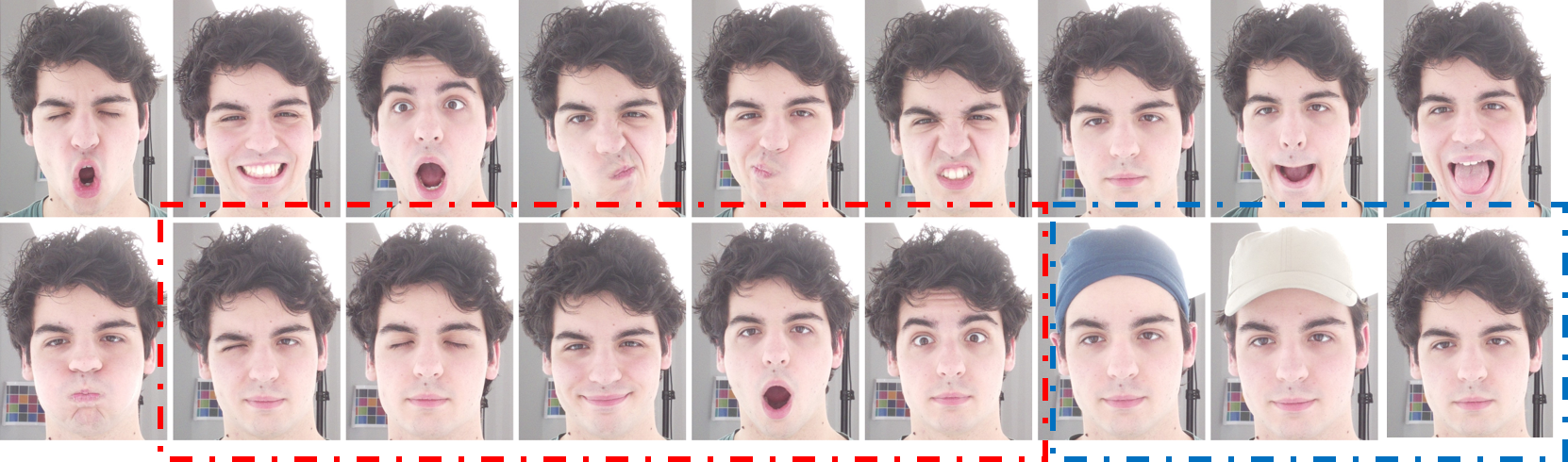}
    \caption{Overview of the captured expressions (red box: test set; blue box: hairstyles) for each subject. 
    }
    \label{fig:expressions} 
\end{figure}
\begin{figure}[t]
{\includegraphics[width=\linewidth]{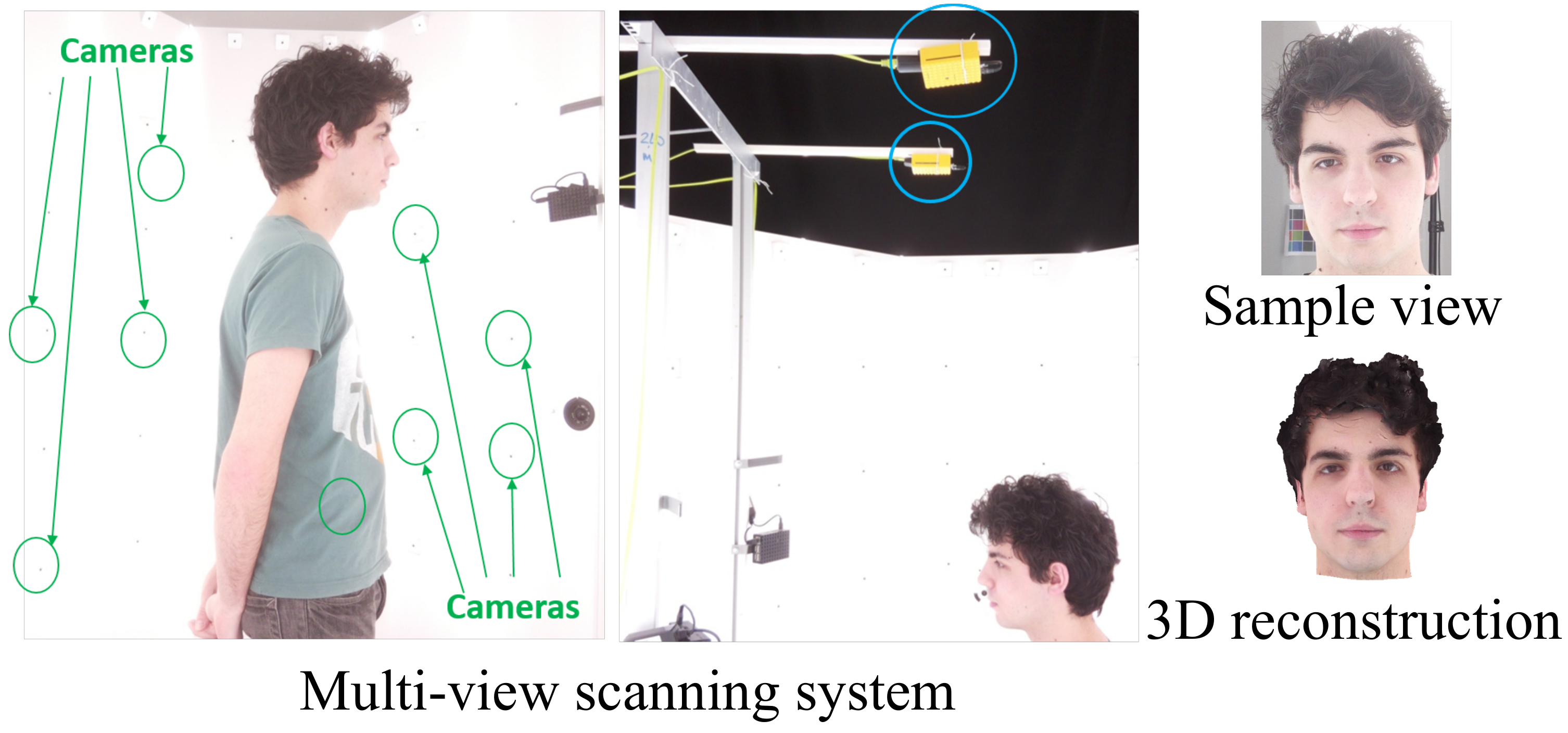}}
    \caption{Multi-view scanning system with $135$ cameras around the person (green), and $2$ cameras (blue) on top. Right: front view image, and reconstruction.}
    \label{fig:MMsys} 
\end{figure}
\subsection{Data Acquisition}
For training, we have scanned \totalIdentities{} subjects ($46$ male, $18$ female; $22$ Caucasian, $9$ Asian, $25$ Indian Sub-continent, $3$ Hispanic, $3$ Middle Eastern, $2$ undisclosed; with an age ranging from $19$ to $69$ with average $26$). For each subject we have recorded 10 facial expressions (a subset of which are chosen from paGAN~\cite{nagano2018pagan}), including neutral expressions with 3-4 different ``hairstyles'' (open hair, tied hair for subjects with long hair, and two different caps), as shown in Fig.~\ref{fig:expressions}. 
Our data was acquired with the \emph{Treedys}\footnote{https://www.treedys.com/} multi-view scanning system, that comprises of $137$ calibrated cameras, see Fig.~\ref{fig:MMsys}.
The total capture time per person amounts to about $20$ minutes. 
Photogrammetry-based 3D reconstruction on the multi-view data was applied to obtain the final textured meshes (see Fig.~\ref{fig:MMsys}), where each mesh comprises of around $50{,}000$ vertices.
\subsection{Training Data} \label{sec:trainingdata}
Given our recorded textured mesh data, we apply several preprocessing steps to make the data is suitable for head model learning. This includes a semi-automated landmark annotation,  cropping of the head area to discard the shoulders and parts of the upper body, rigidly aligning  the meshes, and closing the hole at the neck to make the mesh watertight. In the following we elaborate on these steps.

\textbf{Landmark Annotation.} First, we apply the automated face landmark detector from \emph{dlib}~\cite{dlib09} on the front-facing image obtained from the multi-view camera acquisition setup, which produces a total of $66$ face landmarks. We select a subset of 8 landmarks, i.e. the four corner points of the eyes, the tip of the nose, the two corner points of the mouth, and the chin, and then  transfer these 2D landmarks to the 3D mesh.
We manually correct the inaccurately detected landmarks, and additionally annotate $8$ landmarks for the corners of ears (top, bottom, left, and right) directly in 3D.

\textbf{Head Cropping.} In order to remove the shoulders and parts of the upper body, we crop the head meshes based on the 3D landmarks. To this end, we first compute the vector $\mathbf{v}$ from the centroid of the four eye cornerpoints to the tip of the nose. Then, for $\mathbf{p}$ being the chin landmark in 3D space, we define a virtual plane that goes through the point $\mathbf{p} + \frac{1}{2} \mathbf{v}$ and has the normal $\mathbf{v}$. We only keep the side of the plane that contains the facial landmarks.

\textbf{Rigid Alignment.} 
Based on the 3D facial landmarks, we perform a rigid alignment of all head scans in order to ensure that they are oriented consistently in 3D space. We use a numerically stable implementation~\cite{bernard2015transitively} of the transformation synchronization method~\cite{Bernard_2015_CVPR}, which solves the generalized Procrustes problem in an initialization-free and unbiased way. To this end, we first rigidly align all pairs of heads based on Singular Value Decomposition (SVD),~i.e.~we solve all pairwise Procrustes problems, and subsequently use transformation synchronization to establish cycle-consistency in the set of pairwise transformations. 

\textbf{Hole Closing.} Since our full head model is based on a signed distance representation, we require that the meshes are watertight. After rigid alignment, we assume that the longitudinal axis of the head aligns with the y-axis. With that, we use a flat patch to close the hole at the neck by extruding the respective boundary vertices to coincide with the smallest y-coordinate. Finally, we scale all meshes with the same value such that all of them fit in the unit cube.
\subsection{Training}
\label{sec:training}
We learn a vector-valued function $f_{\theta}(\mathbf{x},\mathbf{z})$, where $\theta$ are the neural network weights,  $\mathbf{x} \in \mathbb{R}^3$ is the query point, and $\mathbf{z}$ is a code vector that encodes the head instance.
The output $f_{\theta}(\mathbf{x},\mathbf{z}) = ({s}, \mathbf{c})$ includes a scalar ${s}$ that represents the signed distance to the surface, as well as the color $\mathbf{c} \in \mathbb{R}^3$ at the closest surface point from $\mathbf{x}$.
 The shape boundary is represented as the zero level-set of the signed distance function (SDF), while the interior parts of the shapes have a negative signed distance value, and the exterior parts a positive value.
We use an autodecoder network architecture~\cite{Park_2019_CVPR}, where the weights of the network $\theta$ and the input latent codes $\mathbf{z}$ for all shapes are learned jointly. 

\textbf{Mesh Sampling.} We require $(\mathbf{x}, 
{s}, \mathbf{c})$-triplets (query point, signed distance value, color) for training. We use a combination of two strategies for sampling these triplets. First, we sample points on the mesh surface based on uniform random sampling. However, in order to also account for higher accuracy in high-detail facial regions, i.e. the eyes, nose and mouth, we additionally sample more points in these areas. We take the center of each eye, the tip of the nose, and the center of the mouth, and place a sphere around them that covers the respective region of interest. Then, we sample points on the mesh surface that lie within each sphere. Eventually we mix the uniformly sampled points with the landmark-based sampled points in a $1$ to $3$ ratio.

After sampling, the points are perturbed by a uniform 3D Gaussian with standard deviation $0.005$ times the length of the bounding box, such that not only surface points but also points in the interior and exterior are used, see~\cite{Park_2019_CVPR}.
The color value for each sample is obtained by finding the closest point on the mesh surface, and then looking up its color in the texture map.%

\textbf{Latent Codes and Disentanglement.}
As mentioned earlier, latent codes for each object (head scan) are also learned during training. 
Existing approaches use a single object latent code to describe the shape. 
In contrast, we design several separate latent spaces for our objects in order to learn a semantically disentangled model.
We use two separate latent vector spaces for geometry and color, $\mathbf{z}_\text{geo}$ and $\mathbf{z}_\text{col}$, respectively. 
The geometry space includes three code vectors for identity, expression, and hairstyle, and the color space includes two code vectors for identity and hairstyle.
Thus, $\mathbf{z}_\text{geo} =  (\mathbf{z}_\text{geoId}, \mathbf{z}_\text{geoEx}, \mathbf{z}_\text{geoH})$ and $\mathbf{z}_\text{col} =  (\mathbf{z}_\text{colId}, \mathbf{z}_\text{colH})$. 
During training, the number of different identity code vectors is equal to the number of training identities, \trainingIdentities{}.
The number of different expression vectors is fixed to $10$ (cf. Fig.~\ref{fig:expressions} for the training expressions) and hairstyle to $4$ (\emph{short}, \emph{long}, \emph{cap1} or \emph{cap2}) for geometry, and to $3$ (\emph{nocap},  \emph{cap1} or \emph{cap2}) for color.
For all scans of the $i$-th subject we use the same latent variables for the geometry identity and color identity, $\mathbf{z}_\text{geoId}^i$ and $\mathbf{z}_\text{colId}^i$.
Similarly, for each expression and hairstyle, the same variables $\mathbf{z}_\text{geoEx}, \mathbf{z}_\text{geoH}$ and $\mathbf{z}_\text{colH}$ are used across all identities.
By doing so, we are able to learn disentangled latent variables without imposing any explicit constraints.
At test time, we can control each latent space individually, leading to semantically meaningful editing.  

\textbf{Network Architecture.}
\begin{figure*}[t]
    \includegraphics[width=\linewidth]{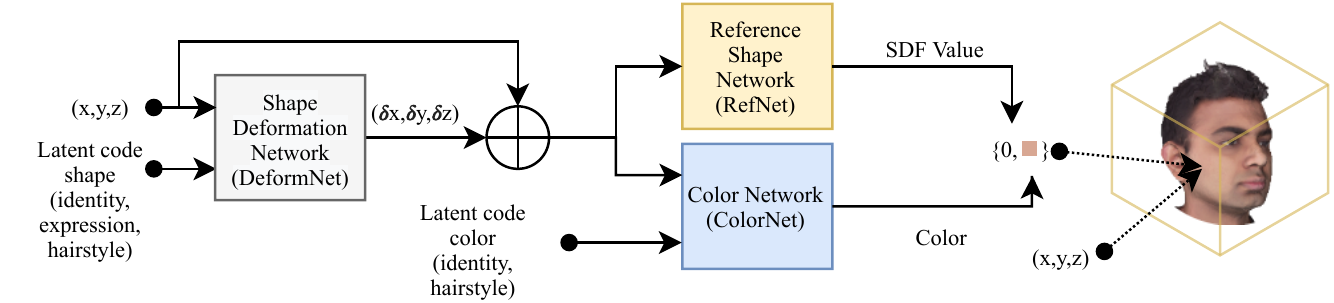}
    \caption{Overview of our network architecture. We learn weights of three network components, a \emph{Shape Deformation} component, a \emph{Reference Shape} component, and a \emph{Color} component. Moreover, the latent codes for each object are also optimized for. The input of the network is a 3D query point, and the output is a signed distance value along with the corresponding color.}
    \label{fig:network} 
\end{figure*}
Our network comprises three components, the \emph{Reference Shape Network} (\emph{RefNet}), the \emph{Shape Deformation Network} (\emph{DeformNet}) and the \emph{Color Network} (\emph{ColorNet}). 
All networks are composed of fully-connected layers with a \emph{ReLU} non-linearity after every layer, except the output layer.
{The inputs to all our networks are encoded using sinusoidal positional encoding~\cite{NeRF}.}

Our \emph{Reference Shape Network} encodes a single reference shape, such that all
individual head shapes can be obtained by deforming this shape. 
The output at a query point $\mathbf{x}$ is the signed distance value $f_\theta^r(\mathbf{x}) \in \mathbb{R}$.
Note that for the reference shape there is no latent code input since only a \emph{single} reference shape is learned. 
This can be seen analogously to the mean (or neutral) shape used in classical 3DMMs~\cite{BlanzVetter}.
We use $3$ fully connected layers for this network, where each hidden layer has dimensionality $512$.

The \emph{Shape Deformation Network} deforms the reference shape to represent the shape of an individual head. 
The network takes the geometry latent code $\mathbf{z}^i_\text{geo}$ for an object $i$, and a query point $\mathbf{x}$ as input, and produces
the output,
\begin{align}
\label{eq:deformdelta}
f_\theta^s(\mathbf{x}, \mathbf{z}^i_\text{geo}) = \delta \in \mathbb{R}^3\,,    
\end{align}
 which is a displacement vector that deforms the query point to the reference shape. 
Thus, the signed distance value at any point $\mathbf{x}$ of the object $i$ with geometry latent $\mathbf{z}^i_\text{geo}$ is 
\begin{align}
    {s}(\mathbf{x}, \mathbf{z}^i_\text{geo}) = f_\theta^r(\mathbf{x} + \delta)
    \,,\label{eq:sdfValue}
\end{align}
where ${s}(\cdot)$ is the scalar signed distance value, and $\delta$ is the deformation from Eq.~\eqref{eq:deformdelta}. 
This formulation allows us to compute dense correspondences between any input head scan and the reference head shape. 
The separation between a reference shape and deformations with respect to this reference shape is also common in classic morphable models~\cite{egger20193d}. 
However, these models require dense correspondences before training, which is not necessary in our approach.
We use $8$ fully connected layers for this network, where each hidden layer has dimensionality $1024$.

The introduction of the reference shape  makes it possible to disentangle the geometry and color components. 
The \emph{Color Network} learns the color of the query point in the reference space.
Given a query point $\mathbf{x}$, deformation $\delta$ from Eq.~\eqref{eq:deformdelta}, and color latent vector $\mathbf{z}^i_\text{col}$ for the object $i$, the output is represented as $f_\theta^c(\mathbf{x}+\delta, \mathbf{z}^i_\text{col}) \in \mathbb{R}^3$, which is the color at point $\mathbf{x}$.
Note that without this separation of a reference space, the \emph{ColorNet} would also have to take into account information about object geometry, thus not being able to disentangle shape and color.
{Note that the latent code for \emph{ColorNet}, for a given identity and hairstyle, does not change with expressions. \emph{DeformNet} finds the right colors and geometry for different expressions by achieving dense correspondences.
We use $9$ fully connected layers for this network, where each hidden layer has dimensionality $1024$.}

\textbf{Loss Functions.}
We define the loss function to train our network as
\begin{align}
    \mathcal{L}_\theta(\mathbf{x}, \mathbf{z_\text{geo}}, \mathbf{z_\text{col}}) & =  \sum_{i=1}^K(\mathcal{L}_\theta^\text{geo}(\mathbf{x}, \mathbf{z}^i_\text{geo})  + 
    \mathcal{L}_\theta^\text{def}(\mathbf{x}, \mathbf{z}^i_\text{geo})  \nonumber \\ &+
    \mathcal{L}_\theta^\text{col}(\mathbf{x}, \mathbf{z}^i_\text{col})  + \mathcal{L}^\text{reg}(\mathbf{z}^i_\text{geo}, \mathbf{z}^i_\text{col}) ) \nonumber \\ &+ \sum_{i\neq j}\mathcal{L}^\text{lm}_\theta(\mathbf{z}^i_\text{geo},\mathbf{z}^j_\text{geo})
    \,,
\end{align}
 where, $i,\,j = \{1,\dots,K\}$, $K$ is the number of scans in the batch, $\mathbf{z}_\text{geo} = \{\mathbf{z}^1_\text{geo},\dots,\mathbf{z}^K_\text{geo}\}$, and $\mathbf{z}_\text{col} = \{\mathbf{z}^1_\text{col},\dots,\mathbf{z}^K_\text{col}\}$.
  The latent vectors of head $i$ are represented as $\mathbf{z}_\text{geo}^i$ and  $\mathbf{z}_\text{col}^i$.
 Here, $\mathcal{L}_\theta^\text{geo}(\cdot)$ enforces good geometry reconstructions, $\mathcal{L}_\theta^\text{def}(\cdot)$ regularizes the deformation field, $\mathcal{L}_\theta^\text{col}(\cdot)$ is used to train the \emph{ColorNet}, $\mathcal{L}^\text{reg}(\cdot)$ is a regularizer on the latent vectors, and $\mathcal{L}^\text{lm}_\theta(\cdot)$ is a sparse pairwise landmark supervision loss.
For the geometry term that we impose upon the signed distance values, 
we use the $\ell_1$-loss
\begin{align}
    \mathcal{L}_\theta^\text{geo}(\mathbf{x}, \mathbf{z}^i_\text{geo}) = w_g \big\| \text{cl}(s(\mathbf{x}, \mathbf{z}^i_\text{geo}), t) - \text{cl}(s_\text{gt}(\mathbf{x}), t) \big\|_1 
    \,{,}
\end{align}
where $s_\text{gt}(\mathbf{x})$ is the ground truth signed distance value at $\mathbf{x}$, and $s(.)$ is from Eq.~\eqref{eq:sdfValue}. 
These values are (symmetrically) clamped at $t$=$0.1$, for which we define $\text{cl}(x, t) := \min(t, \max(x, -t))$. 
We use a similar $\ell_1$-loss for the color component,~i.e.
\begin{align}
    \mathcal{L}_\theta^\text{col}(\mathbf{x}, \mathbf{z}^i_\text{col}) = w_c \big\| f_\theta^c(\mathbf{x}+\delta, \mathbf{z}^i_\text{col}) - c_\text{gt}(\mathbf{x}) \big\|_1
    \,{,}
\end{align}
where $c_\text{gt} (\mathbf{x})$ is the ground truth color at $\mathbf{x}$.
We also enforce that the $16$ landmarks $\{\mathbf{x}^{\ell}_i\}$, as described in Sec.~\ref{sec:trainingdata}, of each shape $i$, deform to the same points in the reference space using the pairwise loss
\begin{align} \label{eq:pairwiseloss}
    \mathcal{L}^\text{lm}_\theta(\mathbf{z}^i_\text{geo},\mathbf{z}^j_\text{geo}) = w_{lm} \sum_{\ell=1}^{16} \big\| (\mathbf{x}^\ell_i + \delta^\ell_i) - (\mathbf{x}^\ell_j + \delta^\ell_j)  \big\|_2\,,
\end{align}
where, $\delta^\ell_i = f_\theta^s(\mathbf{x}^\ell_i, \mathbf{z}^i_\text{geo})$ as in Eq.~\eqref{eq:deformdelta}.
For scans with the ears covered by hair, we do not have any ear annotations. 
We would like to compress the ear region in the reference shape to a single point in the reconstruction for these shapes. 
Thus, we additionally optimize for one point for each ear. We enforce pairwise constraints between the learnable ear points and the annotated ear points for the other shapes in the batch using Eq.~\eqref{eq:pairwiseloss}.

Further, to ensure regularized deformations to the reference shape, we impose a loss on the amount of deformation. We use $\mathcal{L}_\theta^\text{def}(\mathbf{x}, \mathbf{z}^i_\text{geo}) = w_s || f_\theta^s(\mathbf{x}, \mathbf{z}^i_\text{geo}) ||_2$.
Finally, we use an $\ell_2$-regularizer on the latent vectors assuming a Gaussian prior distribution,
$\mathcal{L}^\text{reg}(\mathbf{z}^i_\text{geo}, \mathbf{z}^i_\text{col}) =w_r ( ||\mathbf{z}^i_\text{geo}||_2 + ||\mathbf{z}^i_\text{col}||_2)$. 

\textbf{Optimization.}
Given $N$ batches with $K$ objects per batch, we optimize for the network weights and the latent vectors by solving the optimization problem 
\begin{align}
    \argmin_{\theta, \{\mathbf{z}^b_\text{geo}, \mathbf{z}^b_\text{col}\}_{b=1}^N}~~ \sum_{b=1}^N\sum_{\mathbf{x}} \mathcal{L}_\theta (\mathbf{x}, \mathbf{z}^b_\text{geo}, \mathbf{z}^b_\text{col})
    \,,
    \label{eq:opt}
\end{align}
where, the inner sum takes into account all sampled points, as explained above, and we abuse the notation $\mathbf{z}^b_\text{geo}, \mathbf{z}^b_\text{col}$ to now represent latent codes of all the scans in batch $b$.

\section{Experiments}
In this section, we present an experimental evaluation of our head model. We demonstrate that the model can be used for reconstructing scan data.
We present an ablation study to carefully analyze our design choices, and compare i3DMM to state-of-the-art head models. 
We also show dense correspondence results and applications of our model.
Before we present these results, we provide additional information on the neural network training and testing.

\textbf{Training Details.}
We train our networks using PyTorch~\cite{paszke2019pytorch}, where we use the Adam~\cite{kingma2017adam} solver with mini-batches of size $64$. 
We train for $1000$ epochs with a learning rate of $0.0005$, which decays by a factor of $2$ every $250$ epochs. 
We initialize \emph{RefNet} by pretraining it using only one mouth-open (top row, third from left in Fig.~\ref{fig:expressions}) training scan.
Our network takes about $2$ days to train on $2$ NVIDIA RTX8000 GPUs.

\textbf{Test Data.}
We collect a separate test set consisting of scans of \testIdentities{} identities, which are not part of the training data. 
We capture each identity in 5 novel expressions that are not part of the training expressions, see Fig.~\ref{fig:expressions}.
Each scan is preprocessed analogously to the training scans. 
\subsection{Reconstruction}
For a given test scan, we  fit our learned model to it. 
This is done by optimizing for the latent vector that can best reproduce the scan,~i.e.~by finding the latent variables that minimize the problem
\begin{align}
    \argmin_{\mathbf{z}_\text{geo}^i, \mathbf{z}_\text{col}^i}~~ \sum_{\mathbf{x}} &\, (\mathcal{L}_\theta^\text{geo}(\mathbf{x}, \mathbf{z}^i_\text{geo})  + 
    \mathcal{L}_\theta^\text{def}(\mathbf{x}, \mathbf{z}^i_\text{geo})  \nonumber \\ + & 
    \mathcal{L}_\theta^\text{col}(\mathbf{x}, \mathbf{z}^i_\text{col})  + \mathcal{L}^\text{reg}(\mathbf{z}^i_\text{geo}, \mathbf{z}^i_\text{col}))
    \,,
\end{align}
where, $\mathbf{z}_\text{geo}^i, \mathbf{z}_\text{col}^i$ is the latent code for test scan $i$.
This equation is similar to Eq.~\eqref{eq:opt}, with the difference that the network weights are fixed here and the pairwise landmark supervision loss in Eq.~\eqref{eq:pairwiseloss} is not enforced. 
We use Adam~\cite{kingma2017adam} with a step size of $0.0005$  to solve this problem.
We show several reconstruction results on the test data in Fig.~\ref{fig:exp_rec}. 
We can generalize to unseen identities and expressions, even though our training data only consists of \trainingIdentities{} people with $10$  expressions. 
Note that while we can generally preserve the detailed face region of the scans, we also smooth the scans in the noisy hair area.
\subsection{Correspondences}\label{sec:correspondences}
\begin{figure*}[t]
\includegraphics[width=\linewidth]{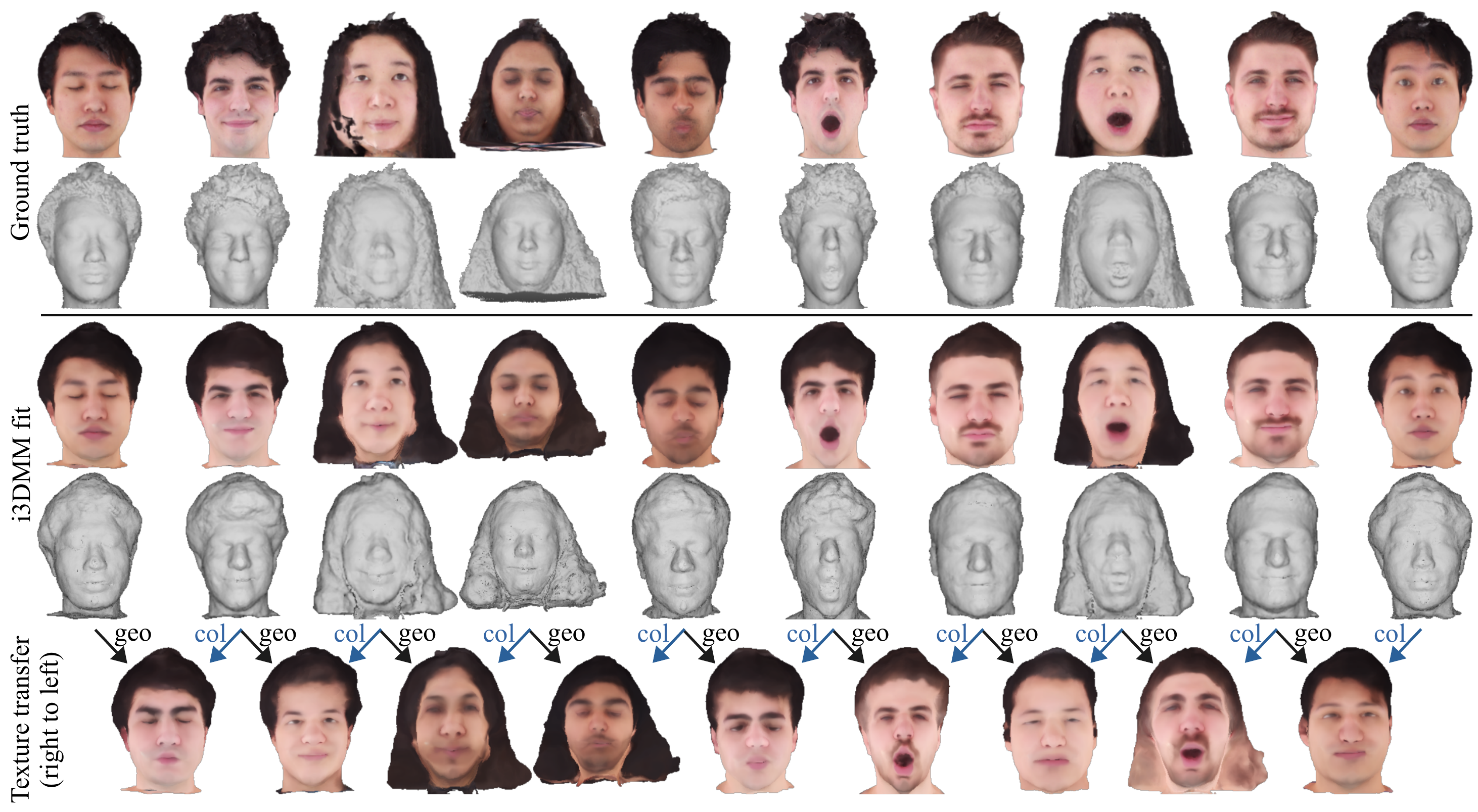}
    \caption{{Reconstruction quality of i3DMM on test data. The top part shows texture and geometry of ground truth scans. The middle part shows texture and geometry of i3DMM fits. The last part shows texture transfer from the scan on right to the scan on left of the image. In column $3$, noise in ground truth does not transfer to the i3DMM fit, showing the robustness of our method.}}
    \label{fig:exp_rec} 
\end{figure*}

{As mentioned in Sec.~\ref{sec:training}, due to the particular design of our network, where the reference shape and the deformations are separated, our approach also establishes dense correspondences between shapes, with extremely sparse landmark supervision. We demonstrate these correspondences in Fig.~\ref{fig:exp_rec}, where the color is transferred from one scan to the other.
The correspondences are also used in the applications of segmentation and landmark transfer in Sec.~\ref{sec:app-seg}.
Our model can reliably find correspondences across different subjects, expressions and hairstyles, including long and short hair. 
Please refer to the supplementary material for more evaluations.} 
\subsection{Ablative Analysis}
\begin{figure}[t]
\includegraphics[width=\linewidth]{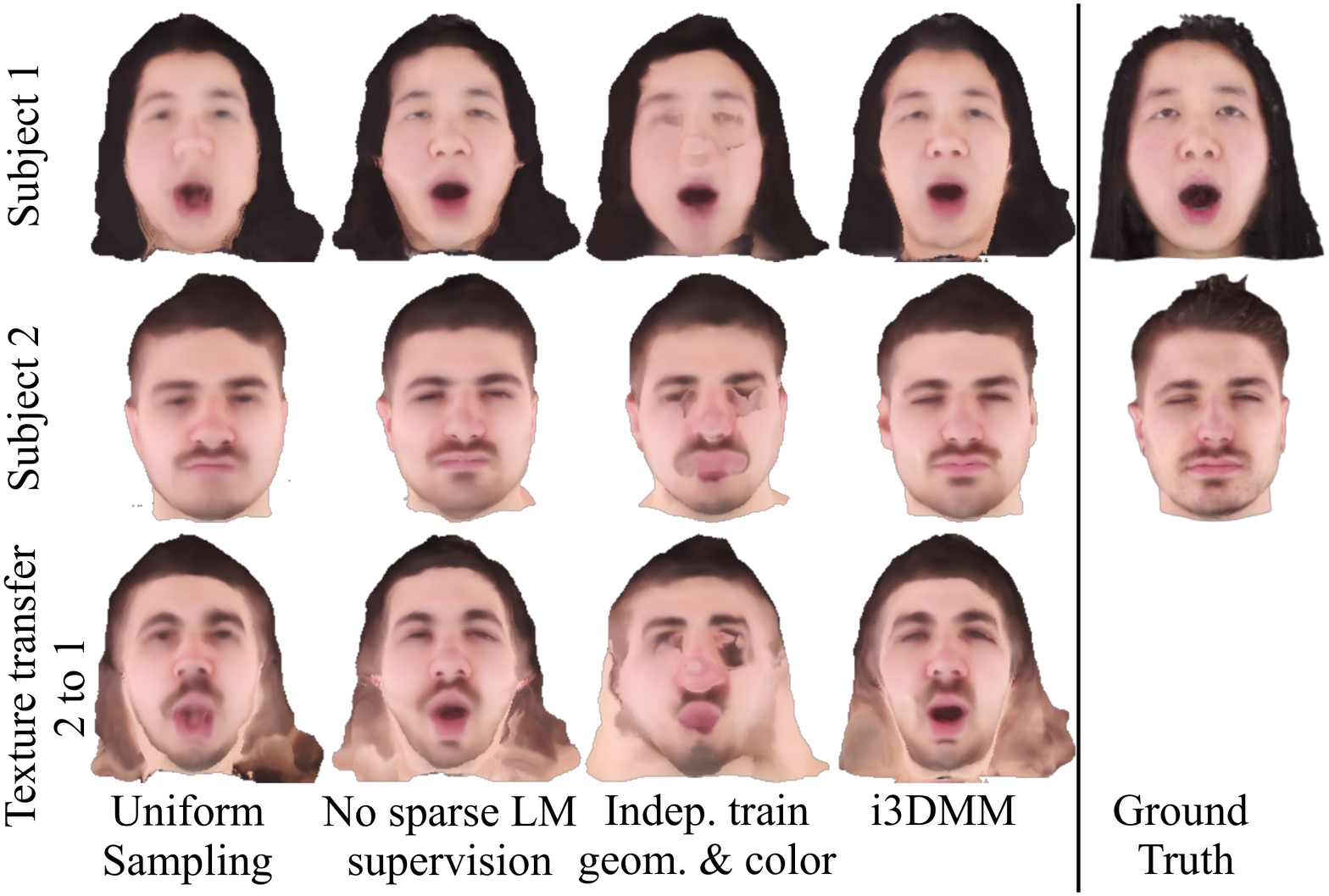}
    \caption{{Ablation study. Top and middle row show reconstructions of two different test scans. Last row shows texture transfer from the scans in the top row to those in the middle row.}
}
\label{fig:exp_abl} 
\end{figure}

{
We design several experiments to evaluate the most important design choices for our approach: landmark-based sampling described in Sec.~\ref{sec:training}, sparse pairwise landmarks loss in Eq.~\eqref{eq:pairwiseloss}, and jointly training \emph{ColorNet}, \emph{DeformNet}, and \emph{RefNet}. We evaluate the qualitative impact of these choices by excluding them one at a time while training our model, see Fig.~\ref{fig:exp_abl}. 
Without landmarks-based sampling, the model focuses more on the noisy hair region and ignores the details in the face region (first column in Fig.~\ref{fig:exp_abl}). Without landmark supervision, the network creates small (fake) ear regions in the hair for the long-haired scans (second column in Fig.~\ref{fig:exp_abl}). It also leads to poor texture transfer around the ear. Finally, for evaluating the importance of joint training of the color and geometry networks, we first train \emph{RefNet} and {DeformNet}. 
After convergence, we train \emph{ColorNet}. 
The correspondences in this case are learned only from geometry reconstructions, which leads to artifacts as shown in the third column of Fig.~\ref{fig:exp_abl}.}
\subsection{Comparisons to Existing Models}
\begin{figure*}[t]
\includegraphics[width=\linewidth]{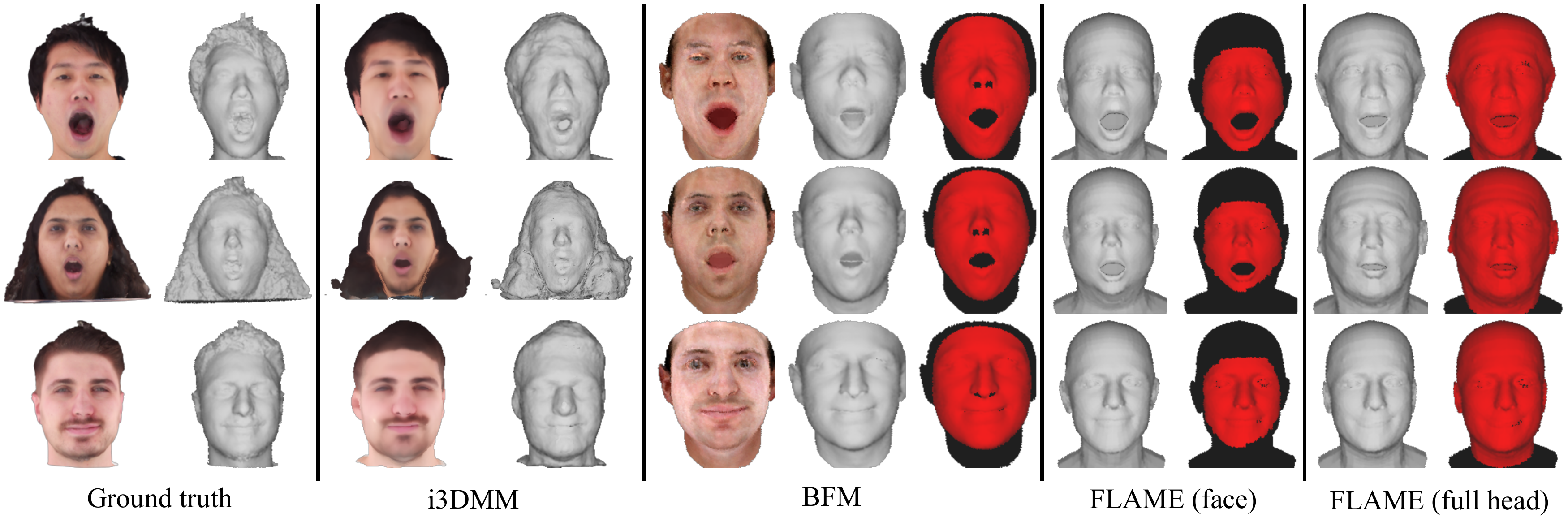}
    \caption{{Comparison of fitting various models to scan data (left). The three considered models are our i3DMM (middle-left), BFM~\cite{Paysan09} (middle), and FLAME~\cite{FLAME:SiggraphAsia2017} (middle-right and right). FLAME does not provide color, and is evaluated in two settings, face-only (middle-right) and full head (right). We visualize both color and geometry for BFM and our results, and only geometry for FLAME. The masks used for fitting are also visualized.}}
    \label{fig:exp_comparison} 
\end{figure*}

We compare our model with the full head FLAME~\cite{FLAME:SiggraphAsia2017}, face only FLAME, and Basel Face Model (BFM)~\cite{Paysan09} which only models the frontal face. %
We fit each model to the test set by optimizing for their model parameters.
Please refer to the supplemental for more details on the fitting.%

We show qualitative results in Fig.~\ref{fig:exp_comparison}. 
FLAME only models the cranium without hair, thus fitting to the full head region results in incorrect head shapes.
We also evaluate FLAME by only using the the face region for fitting.
This leads to higher quality results. 
We obtain higher quality geometry both for the face and  head regions. 
In addition, we can also reconstruct the color of the head, while FLAME can only model the geometry. 
We combine the BFM~\cite{Paysan09} identity geometry and appearance models with the expression model used in Tewari~\etal~\cite{tewari2019fml}, and use this to fit to our test scans.
The expression model is a combination of two blendshape models~\cite{Alexander2009,Cao2014b}.
Since BFM also includes color, we jointly optimize to minimize the geometry as well as color alignment errors.
As can be seen in Fig.~\ref{fig:exp_comparison} (middle), this model can fit to the expression as well as color of the scans. 
However, it is limited to only the face region, while i3DMM can reconstruct the full head. 
In addition, our reconstructions are more personalized, with higher quality nose geometry and face colors.

We show the quantitative evaluations of the different models in terms of the symmetric Chamfer distance and F-score metrics {in Table~\ref{table:models}.
Chamfer distance is the mean distance of points from one shape to their closest points in another. 
We compute the symmetric Chamfer distance as the mean of Chamfer distance from the ground truth to the fits and the Chamfer distance in the opposite direction. 
F-score is ($100$x) the harmonic mean of these two distances after applying a threshold. 
We use a similar metric for color. 
First, the mean error in color at points in the ground truth and the color at their nearest points in the fits is computed, along with the mean error in the opposite direction
Finally, the color error is reported as their average. 
Please refer to the supplemental for details on the metrics used. 
We sample $150{,}000$ points for computing the metrics. 
We achieve similar geometric quality as FLAME and BFM in the face region, but significantly outperform FLAME for the full head reconstructions, see Table~\ref{table:models}.
Moreover, in terms of color, our reconstructions are more realistic than BFM and also outperform it quantitatively.}
\begin{table}
\begin{tabular}{  l|  c | c || c | c | c }
Region& \multicolumn{2}{c||}{\textbf{Full Head}} & \multicolumn{3}{c}{\textbf{Face}}\\ \hline
Metric  & Ours & FLAME  & Ours & BFM & FLAME \\ \hline
Chamfer $\downarrow$ & \textbf{3.31} & 7.83  & 1.02& 0.96& \textbf{0.88} \\
 (mm) &  &  & &  &  \\ \hline
F-score $\uparrow$ & \textbf{63.38} & 45.96 & 99.31 & 97.66& \textbf{99.6}\\ \hline
Color $\downarrow$ & 0.11 & -- & \textbf{0.07} & 0.09 &  --  \\
\end{tabular}
\vspace{2mm}
\caption{Quantitative comparison of head models. Symmetric Chamfer distance is reported here. F-score is computed with a threshold of $0.01$.}
\label{table:models}
\end{table}
\subsection{Application: Semantic Head Editing}
{Due to the semantic disentanglement in our model, we can selectively change semantic components of a head. 
In Fig.~\ref{fig:teaser}, we edit one feature of a test scan  while keeping the other features fixed.
To obtain the edited latent codes, we first find the principle components of variation in each latent-subspace by running PCA on the training latent spaces.
We move along the principal components to pick the latent codes to edit the identity, and we pick others from the trained latent codes.
Unlike existing models, we can also edit hairstyles and add caps while keeping other components fixed. 
We also model the mouth interior. 
These features allow for semantic head editing applications much beyond the capabilities of the existing methods. 
}
\subsection{Application: One-shot Annotation Transfer}
\label{sec:app-seg}
\begin{figure}[t]
\includegraphics[width=\linewidth]{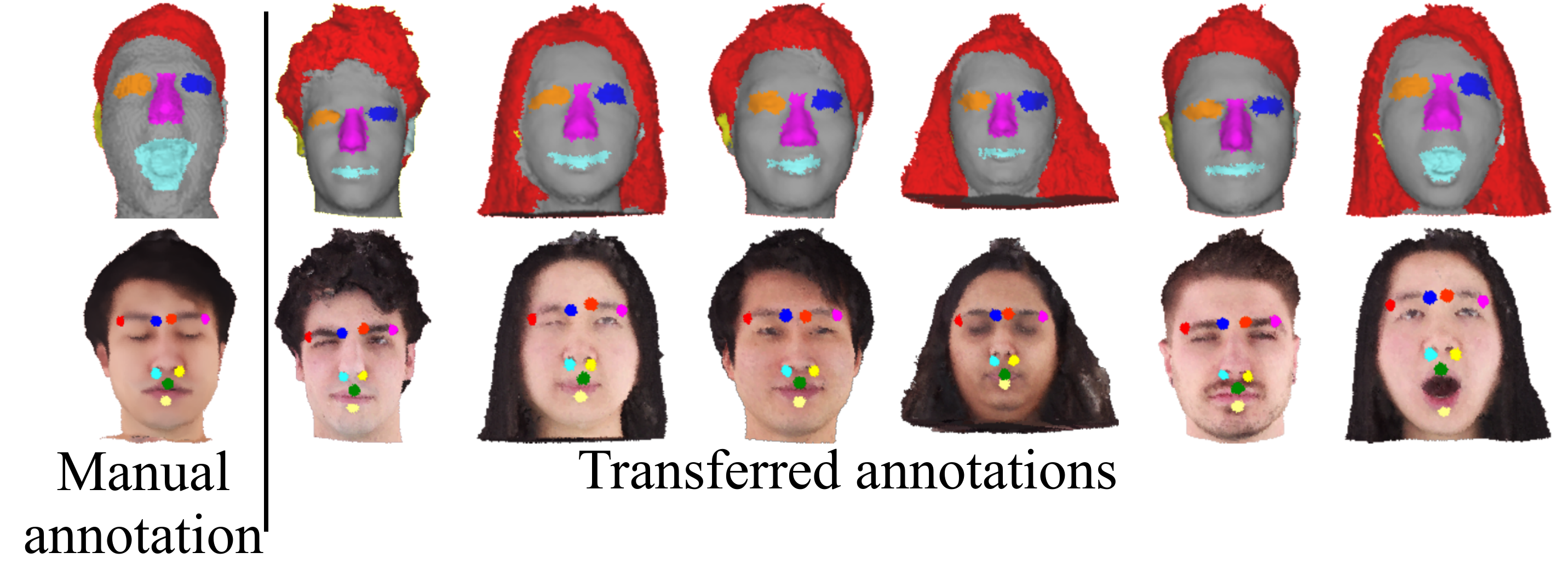}
    \caption{Application: One-shot  annotations on real-world scans. Coarsely drawn annotations (top row: semantic segmentation, bottom row: landmarks, not used during training) on one scan (left) can be automatically transferred to other real-world head scans (right).}
    \label{fig:segmentationTransfer} 
\end{figure}

As i3DMM can predict dense correspondences among head scans, it also enables us to transfer annotations across different reconstructions.
As our model has low reconstruction errors, it can also be used to annotate real-world 3D scans.
We show this application in Fig.~\ref{fig:segmentationTransfer}.
We first transfer the annotations from one manually annotated scan to the reference shape.
We can then transfer them to different i3DMM fits, and also to the real-world scans using nearest neighbors from the fits to the scans.
As i3DMM can be used to even annotate the hair region, it can be very useful to curate scan datasets.
Please refer to the supplemental for more results. 

\section{Discussion and Future Work}
{Although i3DMM disentangles and models an unprecedented amount of variety in head features, and enables reconstructions with very high quality (Fig.~\ref{fig:exp_rec}), it still has some limitations. %
Although i3DMM models varying hairstyle shapes, it is only a coarse approximation of the physical nature of hair comprising of individual hair strands.
Further, we model hairstyles that cover the ear by trying to collapse the ear to a single point (Eq.~\eqref{eq:pairwiseloss}).
Ideally, the layered geometry of ears behind the hair should be correctly represented. 
Finally, there is room for extending i3DMM for more fine-grained control over hairstyles. 
One challenge is that our scans are noisy in the hair region. 
This is due to the photometric multi-view reconstruction techniques used, which struggle with hair due to their complex physical interaction with light.
}

\section{Conclusion}
We have presented the first unified 3D morphable model of geometry and appearance of full heads, which includes different identities, expressions and hairstyles. 
The head geometry and color is represented using implicit function parameterized with neural networks, learned from $610$ multi-view scans.
By explicitly separating the network into reference shape, shape deformation and color components, our model can disentangle the different semantic components.
It also allows us to obtain dense correspondences across different scans represented using our model, enabling several applications. 
We will make our code publicly available to the community. 

\section{Acknowledgements}
This work has been supported by the ERC
Consolidator Grant 4DReply (770784), and by the BMBF-funded Munich Center for Machine Learning. We thank Garvita Tiwari, Navami Kairanda, and Neng Qian for their support in capturing the dataset. We thank all the subjects in the dataset for lending their data for research.
{\small
\bibliographystyle{ieee_fullname}
\bibliography{egbib}
}
\clearpage

\begin{figure*}[t]
    \includegraphics[width=\linewidth]{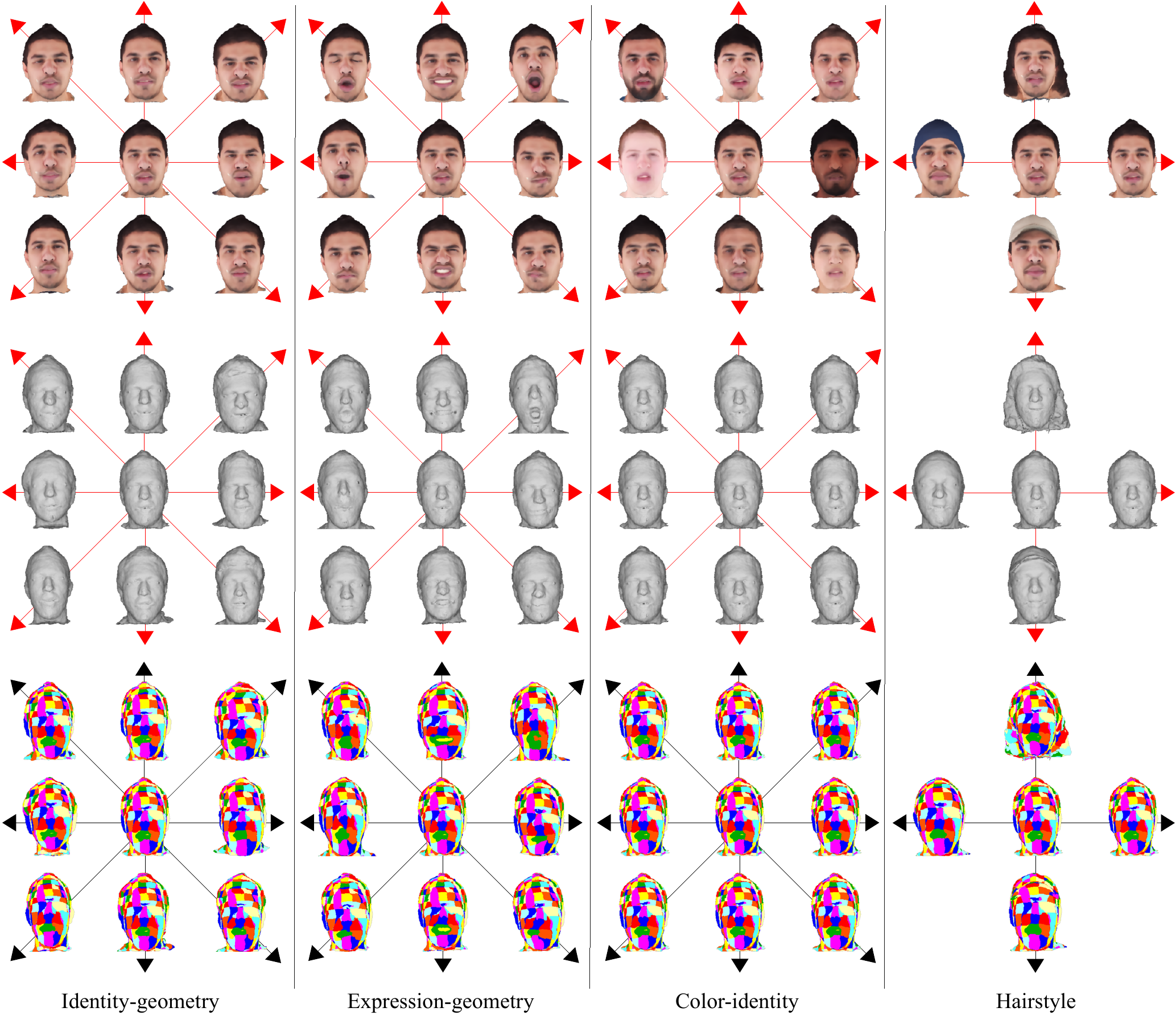}
    \captionof{figure}{Principal components of different spaces i3DMM models. We show color renders at the top, geometry renders in the middle, and correspondences at the bottom.}
    \label{fig:spaces}
\end{figure*}
\section{Supplementary}
\subsection{Model Visualization}
Our implicit 3D morphable model models the identity, expression, and hairstyle components of the geometry; as well as the identity and hairstyle components of the color of the head with independent parameteric controls.
We visualize these components in Fig.~\ref{fig:spaces}.
We perform PCA on the identity geometry and color spaces in order to compute the principal components. 
The expression component is visualized by moving along the directions of the training expressions, since they are semantically well defined. 
As we model hairstyles that include caps, we show the joint space of geometry and color for hairstyles.
Note that the hairstyle geometry can only take four discrete values -- short, long, cap1, or cap2.
Any variation within these categories is modelled by the identity-geometry component. 
Similarly, hairstyle color can only take the values -- nocap, cap1, or cap2. 
The color of hair without any cap is determined by the identity-color component.

\subsection{Experiments}
Here, we provide more details on the evaluations in the main paper, and include further evaluations.
\subsubsection{Sampling i3DMM}
One of the important features of a 3D Morphable Model is the ability to randomly sample shapes in the parametric space. 
This has been used for generating synthetic data for training CNNs~\cite{KimZTTRT17,Richardson_2017_CVPR,sela2017unrestricted}. 

We achieve sampling in i3DMM by performing principal component analysis (PCA) over  the training latent codes for color-identity, geometry-identity, and expressions.
We weigh the singular values using a Gaussian random variable $\mathcal{N}(0,0.1)$ for color-identity, and geometry-identity, and $\mathcal{N}(0,0.25)$ for expressions.
Since the latent codes for hair shapes and colors can take very limited numbers of values and are very well defined semantically, we sample these from their training values. 
We show several results in the supplemental video.

As can be seen, our model is biased towards generating male heads. 
This is likely due to our gender-biased training dataset with $46$ males and $18$ females. 
While this does not lead to any clear loss of quality when fitting to female test scans, see Fig. 6 in the main paper, it might lead to biased quality of results in other problems, for eg., if random samples from the model was used for training another network. 
\subsection{Comparisons} \label{sec:suppComps}
As mentioned in Sec. 4.4 of the main paper, we compare our model to two existing models, BFM~\cite{Paysan09} and FLAME~\cite{FLAME:SiggraphAsia2017}. We show more qualitative model fitting results in Fig.~\ref{fig:more_comparisons}.
Next, we provide more details on the fitting algorithm used. 

\textbf{Fitting:} We paint the face region of each model's template mesh to create a mask as shown in Fig.~\ref{fig:more_comparisons}.
We use these masked regions of the models for fitting the models to head scans.
To initialize, we mark $8$ landmarks (eye corners, nose, lip corners, and chin) on the template mesh and rigidly align the template to each ground truth scan using Procrustes algorithm.
We allow for translation, rotation, and scaling. We use the rigid alignment as initialization and optimize for the parameters of each model using a modified iterative closest point (ICP) algorithm which also updates the model parameters.
The fitting algorithm maintains the initial scale and translation but optimizes for rotation.
In each optimization step, we first compute the correspondences as the closest points from the masked region in the model, shown in Fig.~\ref{fig:more_comparisons}, to the scan data.
We compute the loss as shown in Eq.~\eqref{eq:lossForFitting} and update the model parameters along with Euler angles for global rotation.
We run the following optimization program up to convergence to fit the models to our scans:

\begin{align} \label{eq:lossForFitting}
    \argmin_{\theta,K,\alpha,\beta,\gamma}& \,\, \sum_{i=1}^N \,\,(\big\| sR(\alpha,\beta,\gamma)x_i(\theta) + t - x_i \big\|_2 \nonumber \\ &\quad\quad + \big\| Kc_i(\theta) - c_i \big\|_2 )  \nonumber  \\ &+ w_l \sum_{j=1}^L \big\| sR(\alpha,\beta,\gamma)l_j(\theta) + t - l_j\big\|_2 \,,
\end{align}
where, $x_i(\theta) \in \mathbb{R}^3$ is a vertex $i$ in the masked region of the model (containing $N$ vertices), $x_i$ is the point on the scan data corresponding to $x_i(\theta)$; $R(\alpha,\beta,\gamma) \in \mathbb{R}^{3 \times 3}$ is the global rotation matrix computed using the Euler angles, $\alpha,\beta,\text{and } \gamma$ ($\in \mathbb{R}$); $s  \in \mathbb{R} $, $t \in \mathbb{R}^3$ are the global scale and translation computed during initialization; $c_i(\theta) \in \mathbb{R}^3$, $c_i \in \mathbb{R}^3$ are the colors at the vertices $x_i(\theta)$ and $x_i$ respectively; $l_j \in \mathbb{R}^3$ and $l_j(\theta) \in \mathbb{R}^3$ are the $L(=8)$ ground truth and model landmarks respectively, as described earlier; and $K \in R^{3\times3}$ is a diagonal matrix.
We set $w_l=0.1$ during the fitting process.

Note that the color loss is only enforced for BFM, as FLAME does not model colors. Further, as the color intensities of BFM and our scans differ, we globally scale the color values using channel-specific scalars arranged as a diagonal matrix $K$ which we optimize for along with the model parameters.

\textbf{Evaluation details:}

We describe the evaluation metrics in Sec. 4.4 of the main paper. Here, we present details about the masks used to evaluate these metrics.

\textbf{Face region:} We manually paint face masks on the ground truth scans to obtain the ground truth masks.
We exclude the mouth interior of the ground truth scans.
We copy this mask to the i3DMM fits.
We do that by annotating a vertex in i3DMM reconstruction if the nearest point from that vertex on the ground truth scan is in the masked region.
We show the face masks used to fit BFM and FLAME to ground truth scans in Fig.~\ref{fig:more_comparisons}.
We obtain the symmetric metrics presented in Table. 1 of the main paper for the face region in the following way.
In one direction, we compute the errors from masked region of ground truth to the closest points on the (unmasked) models fit to the scan.
In the other direction, we compute the errors from the masked region of the models to the (unmasked) ground truth scan.
We compute errors between the masked regions of one mesh to unmasked regions of other mesh to avoid large error metrics due to annotation mistakes during manual mask painting.

\textbf{Full Head:} We only fit to FLAME full head model as BFM does not model the entire head.
We remove the neck region from FLAME as shown in Fig.~\ref{fig:more_comparisons} as the ground truth head scans do not have neck regions.
We also remove the vertices used to close the neck from ground truth as FLAME has a hole in the mesh at the neck.
We compute the metrics as we do for the face region between these two full head meshes.
We report the full head metrics for our model in the entire head region, including the closed hole at the neck mesh.
\subsubsection{Ablative Analysis}
In Fig.~\ref{fig:more_ablatives}, we show additional qualitative results for the ablative analysis. 
We also show the quantitative results for full head i3DMM fit in comparison to i3DMM variants. We compare the four models that evaluate our design choices in Table~\ref{tab:ablation}.
The error metrics are computed for the face region using manually annotated face masks as described in Sec.~\ref{sec:suppComps}.
We only evaluate the face region, as the ground truth for hair is noisy, and small quantitative differences are not very indicative of degradation in quality.
Although the geometric reconstruction accuracy is marginally better without the landmark supervision loss, as compared to i3DMM, the color reconstruction accuracy of i3DMM is higher.
Also, as mentioned in the main paper, Sec. 4.3, texture transfer results around the ear regions with landmark supervision loss are worse compared to i3DMM.
\begin{table*}
\begin{center}
\begin{tabular}{ l| c | c | c | c}
& Uniform & No landmark& Independently trained& i3DMM\\ 
& sampling& supervision & color and geometry& \\ \hline
Chamfer (mm) $\downarrow$ & 1.1065 & \textbf{0.9775} & 1.0319& 1.0143\\ \hline
F-score $\uparrow$ & 98.5031 & \textbf{99.5339} & 99.1276& 99.3101\\ \hline
Color $\downarrow$ & 0.0734 & 0.0681& 0.0796& \textbf{0.0655}\\ \hline
\end{tabular}
\caption{Quantitative results for ablation study. The columns, from left to right, show results obtained with  uniform sampling for SDF instead of landmark-based sampling, without sparse pairwise landmark supervision loss, independently training for representing geometry and color, final model (i3DMM), and ground truth.}
\label{tab:ablation} 
\end{center}
\end{table*}
\subsubsection{Correspondence Evaluation}
We quantitatively evaluate the correspondences predicted by i3DMM by using the FLAME and BFM fits as ground truth correspondences.
To this end, we first find the closest points from the vertices of the (masked) model fits to the i3DMM reconstructions for different scans. 
We will call these correspondences ground truth annotations here. 
We use a KD tree algorithm for efficiency. 
The masked face region contains $26370$ vertices for BFM,  and $1873$ vertices for FLAME.
We also transfer the annotations for one i3DMM reconstruction, to all the other reconstructions.
This process is same as that described in annotation transfer application (see Sec. 4.5 of main paper).
We compute the correspondence error as the average of error between the transferred and the ground truth annotations.
Note that we transfer annotations from one i3DMM fit to every other i3DMM fit.
Therefore, we compute a symmetric error metric.

\begin{figure}
\includegraphics[width=\linewidth]{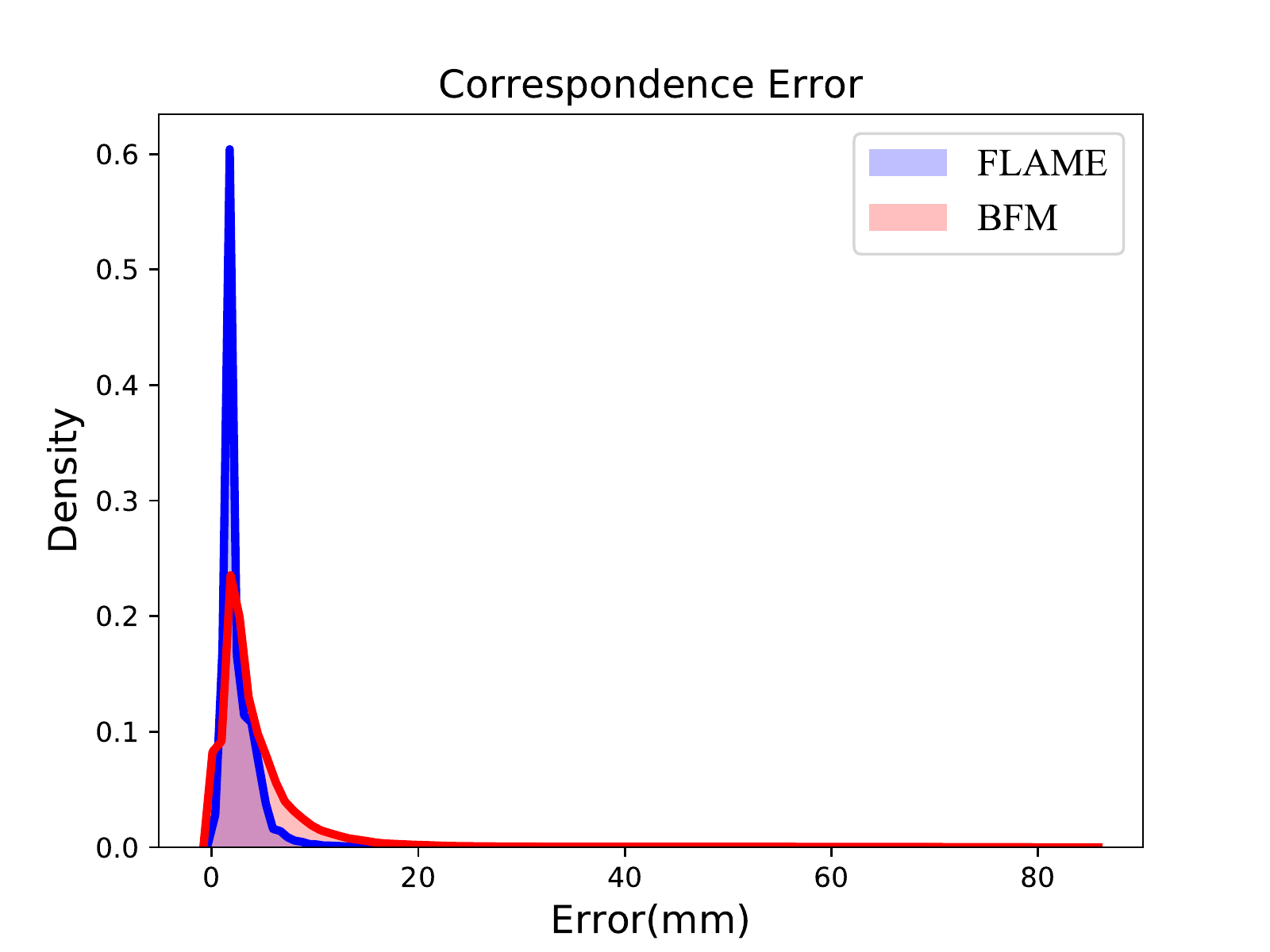}
\caption{Distribution of errors in correspondences predicted by i3DMM computed using FLAME fits (face) as ground truth (blue), and using BFM fits (face) as ground truth (red).}
\label{fig:correspondenceEval} 
\end{figure}

The resulting distribution of error is shown in Fig.~\ref{fig:correspondenceEval} (evaluating with BFM as ground truth is plotted in red, while FLAME is plotted in blue).
The mean and median of errors for BFM is $5.08$mm and $3.02$mm respectively.
The mean and median of errors for FLAME is $2.36$mm and $1.83$mm respectively.
Note that, this error does not only capture the error in i3DMM's correspondence predictions but also the error in registrations of FLAME and BFM fits, see Table. 1 in the main paper.
\subsection{Applications}
\subsubsection{Full Head Completion}
\begin{figure}
\includegraphics[width=\linewidth]{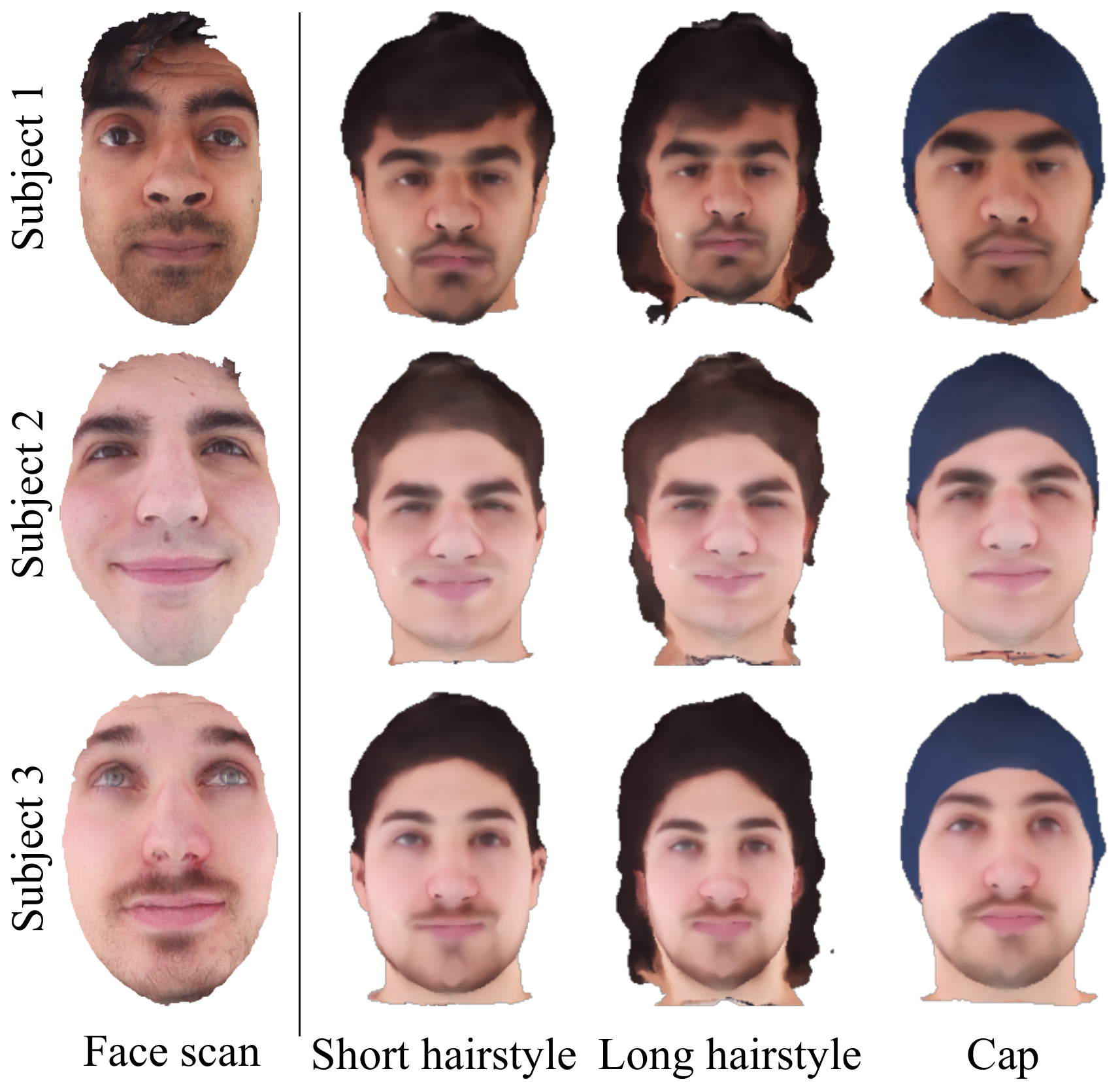}
    \caption{Completing face scans (left) with different hair styles (short hairstyle (middle-left), long hairstyle (middle-right), and cap (right)) using i3DMM as prior. }
    \label{fig:face_completion} 
\end{figure}
We use the i3DMM prior to complete face scans with different hairstyles as shown in Fig.~\ref{fig:face_completion}.
To obtain the face meshes from our head meshes, we delete all the vertices that are outside a sphere around the the tip of the nose.
We learn the latent vector for the given test scan using the SDF samples of the face mesh, as described in Sec. 4.1 of the main paper. 
Additionally, we semantically control the hair style of the completed scan by adding a regularizer that enforces the learned $\mathbf{z}_\text{geoH}$ and $\mathbf{z}_\text{colH}$ to be close to the hairstyle latent vectors learned during training.

It can be inferred from the results in Fig.~\ref{fig:face_completion} that our model learns a good prior distribution, generating plausible heads for the given faces.
i3DMM offers user-guided control for head completion and can be used to turn existing face-only 3DMMs into full head 3DMMs. Further, our method can also be used as a prior distribution for applications such as monocular 3D reconstruction~\cite{tewari17MoFA}. 
\begin{figure*}[t]
\includegraphics[width=\linewidth]{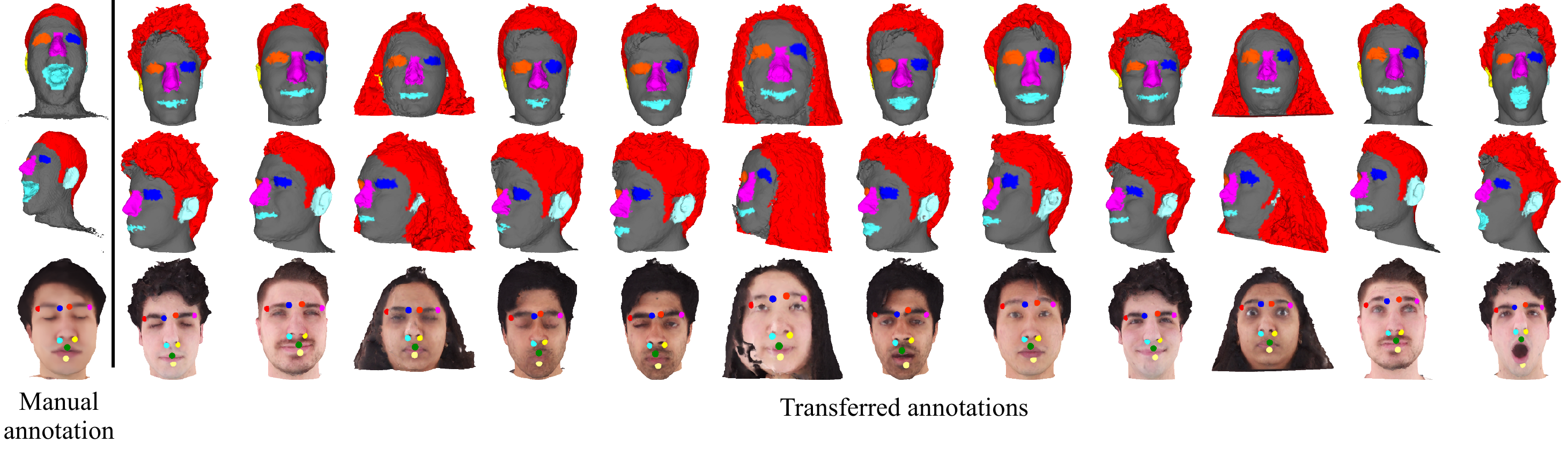}
    \caption{Additional annotation transfer results. Top: segmentation transfer front view. Middle: segmentation transfer side view. Bottom: landmark transfer. Left column shows i3DMM reconstructions with manual annotations. Right part shows annotations transferred to head scans using i3DMM.
     \label{fig:seg_transfer}
}
\end{figure*}
\subsubsection{Annotation Transfer}
We show more results for annotation transfer described in Sec. 4.5 of the main paper, in Fig.~\ref{fig:seg_transfer}.
\begin{figure*}[t]
\includegraphics[width=\linewidth]{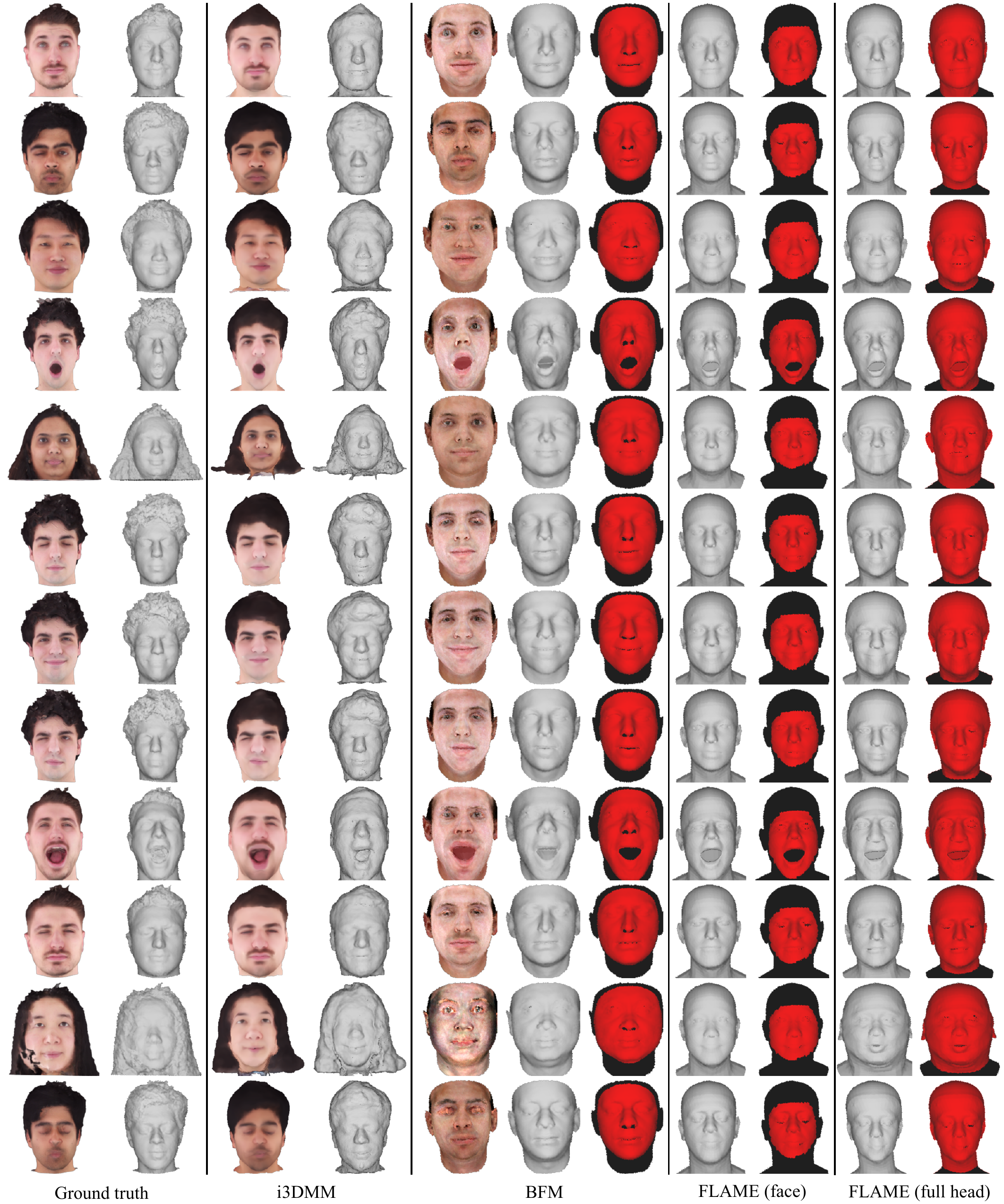}
\caption{Additional comparison results between i3DMM (full head) fits, BFM (face) fit, and FLAME (full head and face) fits.}
\label{fig:more_comparisons} 
\end{figure*}
\begin{figure*}[t]
\includegraphics[width=\linewidth]{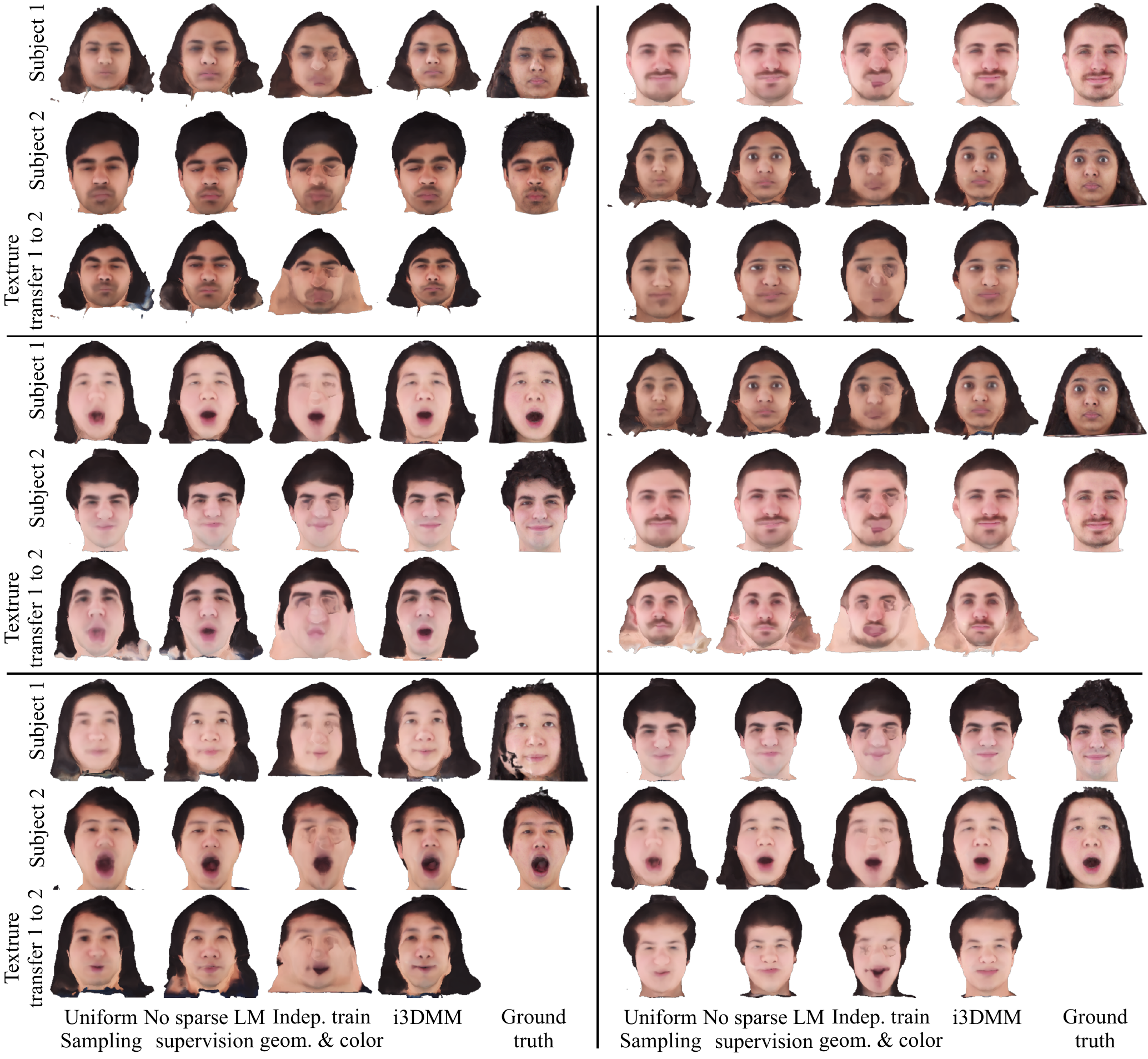}\\
    \caption{Additional ablation results. From left to right, i3DMM without landmark-based sampling, i3DMM without landmark supervision, i3DMM with independent color and geometry training, i3DMM, and ground truth results are shown.
     \label{fig:more_ablatives}
}
\end{figure*}
\subsubsection{Visualization Details}
The output of the i3DMM is a signed distance field. Generally, marching cubes algorithm is used to reconstruct the surface from a SDF.
However, based on the resolution used, marching cubes algorithm introduces unpleasant surface artifacts. %
To avoid these artifacts, we used a sphere tracer to render our results.
We used the Blinn-Phong reflection model to shade our geometry results.
We apply a gamma correction with $\gamma=0.65$ for the color renders.
We use Redner~\cite{redner} for rendering meshes.
\end{document}